\documentclass[10pt,twocolumn,letterpaper]{article} 

\usepackage{avss}
\usepackage{times}
\usepackage{epsfig}
\usepackage{graphicx}
\usepackage{amsmath}
\usepackage{amssymb}

\usepackage{microtype}

\usepackage[pagebackref=true,breaklinks=true,letterpaper=true,colorlinks,bookmarks=false]{hyperref}

\avssfinalcopy 

\def\avssPaperID{0018} 

\ifavssfinal\fi
\begin{document}
	
\title{Adaptive Batch Normalization Networks for Adversarial Robustness}
	
\author{Shao-Yuan Lo \hspace{0.3cm} Vishal M. Patel\\
Johns Hopkins University\\
{\tt\small \{sylo, vpatel36\}@jhu.edu}
}
\maketitle
\thispagestyle{empty}

\let\thefootnote\relax\footnote{Copyright: 979-8-3503-7428-5/24/\$31.00 ©2024 IEEE}

\begin{abstract}
	Deep networks are vulnerable to adversarial examples. Adversarial Training (AT) has been a standard foundation of modern adversarial defense approaches due to its remarkable effectiveness. However, AT is extremely time-consuming, refraining it from wide deployment in practical applications. In this paper, we aim at a non-AT defense: How to design a defense method that gets rid of AT but is still robust against strong adversarial attacks? To answer this question, we resort to adaptive Batch Normalization (BN), inspired by the recent advances in test-time domain adaptation.
	We propose a novel defense accordingly, referred to as the Adaptive Batch Normalization Network (ABNN). ABNN employs a pre-trained substitute model to generate clean BN statistics and sends them to the target model. The target model is exclusively trained on clean data and learns to align the substitute model's BN statistics. Experimental results show that ABNN consistently improves adversarial robustness against both digital and physically realizable attacks on both image and video datasets. Furthermore, ABNN can achieve higher clean data performance and significantly lower training time complexity compared to AT-based approaches.
\end{abstract}

\section{Introduction}
Deep networks have played an essential role in the remarkable success of computer vision. However, deep networks are vulnerable to adversarial examples, which can cause wrong predictions through carefully crafted perturbations \cite{goodfellow2015explaining,szegedy2014intriguing}. Therefore, robust adversarial defense approaches are needed, where Adversarial Training (AT) has been a standard foundation of modern defenses \cite{madry2018towards,zhang2019theoretically}. The AT technique has been proven effective, particularly when facing strong white-box attacks \cite{obfuscated}. Nevertheless, AT has a high computational cost as it involves multi-step adversarial example generation for learning robust features, which refrains it from being widely deployed in real-world applications. Additionally, AT is known for causing lower clean data performance and having limited generalization \cite{lo2021defending,tsipras2018robustness,zhang2019theoretically}.

A few defense categories do not use AT. Image transformation-based approaches \cite{guo2017countering,liao2018defense,xu2017feature} are among the most common non-AT research streams. They employ an image transformation at the pre-processing stage to remove adversarial perturbations. The intuition is to result in a clean input being fed to the target model. However, most of these approaches’ effectiveness is caused by obfuscated gradients, giving a false sense of robustness. Almost all of these attempts have been defeated under the white-box threat model \cite{obfuscated}. To elaborate, if attackers are aware of the presence of the defense, they are able to incorporate the defense into their adversary search or combat obfuscated gradients. Therefore, this defense stream is declining.

\begin{figure}[t!]
	\begin{center}
		\centering
		\includegraphics[width=0.48\textwidth]{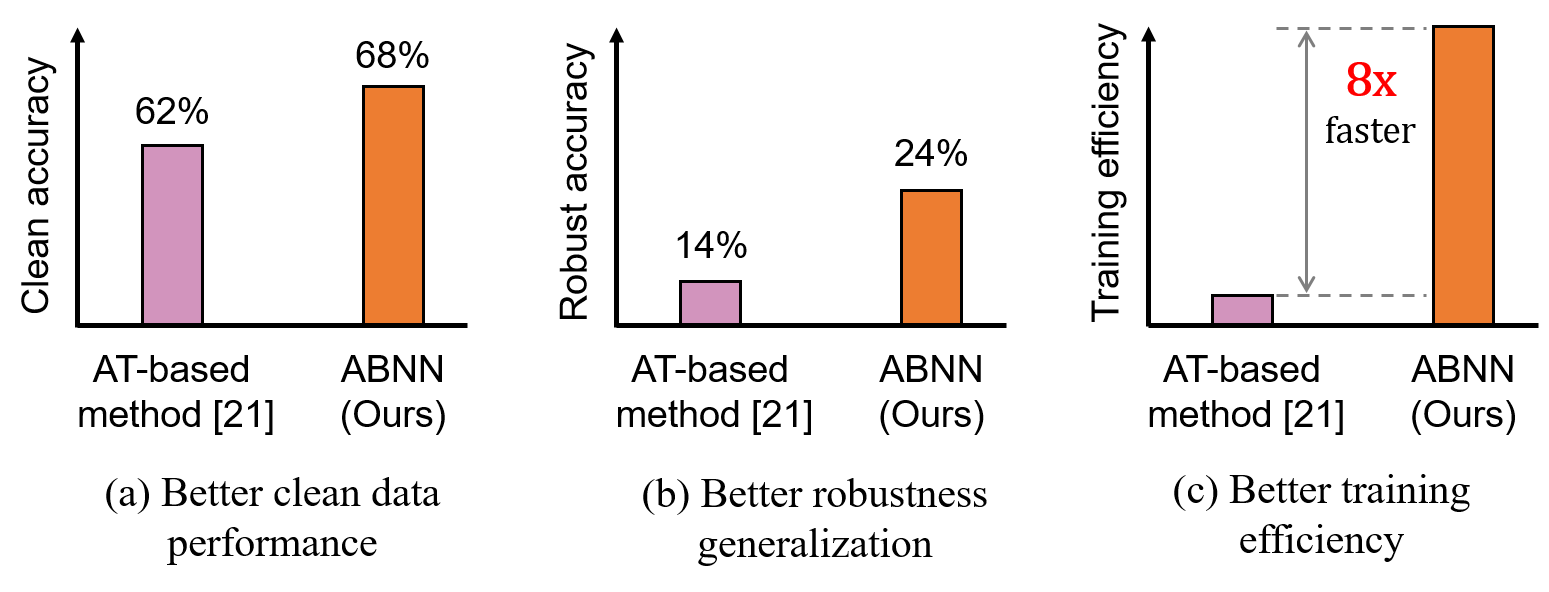}
		\caption{Summary of ABNN's main strengths: (a) Better clean data performance, (b) better robustness generalization, and (c) better training efficiency. Detailed analysis is presented in Section~\ref{sec:3}.}
		\label{fig:cover}
	\end{center}
\end{figure}

In this paper, we aim to develop an effective non-AT defense by addressing the question: \textit{How to design a defense method that gets rid of AT but is still robust against strong adversarial attacks?} To answer this question, we view the adversarial robustness problem from the perspective of domain adaptation \cite{chang2019domain,chang2019all,ganin2015unsupervised,lo2022learning,lo2023spatio}. To be more specific, clean data and adversarial data have distinct distributions \cite{lo2022exploring,lo2021defending,xie2020adversarial,xie2020intriguing}, so we can treat adversarial examples as a kind of domain shift problem. Several studies investigate adversarial effects through domain adaptation techniques. For example, AdvProp \cite{xie2020adversarial} employs an auxiliary Batch Normalization (BN) branch to learn separate clean and adversarial feature distributions, improving image recognition. This idea is originally from the domain adaptation field \cite{chang2019domain,lo2022learning}. Similarly, DRRDN \cite{yang2021adversarial} disentangles the clean and adversarial distributions to enhance robustness, which is also motivated by domain adaptation \cite{chang2019all}. Nevertheless, these works aim at different problem settings from ours and still involve AT.

\begin{figure*}[t!]
	\begin{center}
		\centering
		\includegraphics[width=0.9\textwidth]{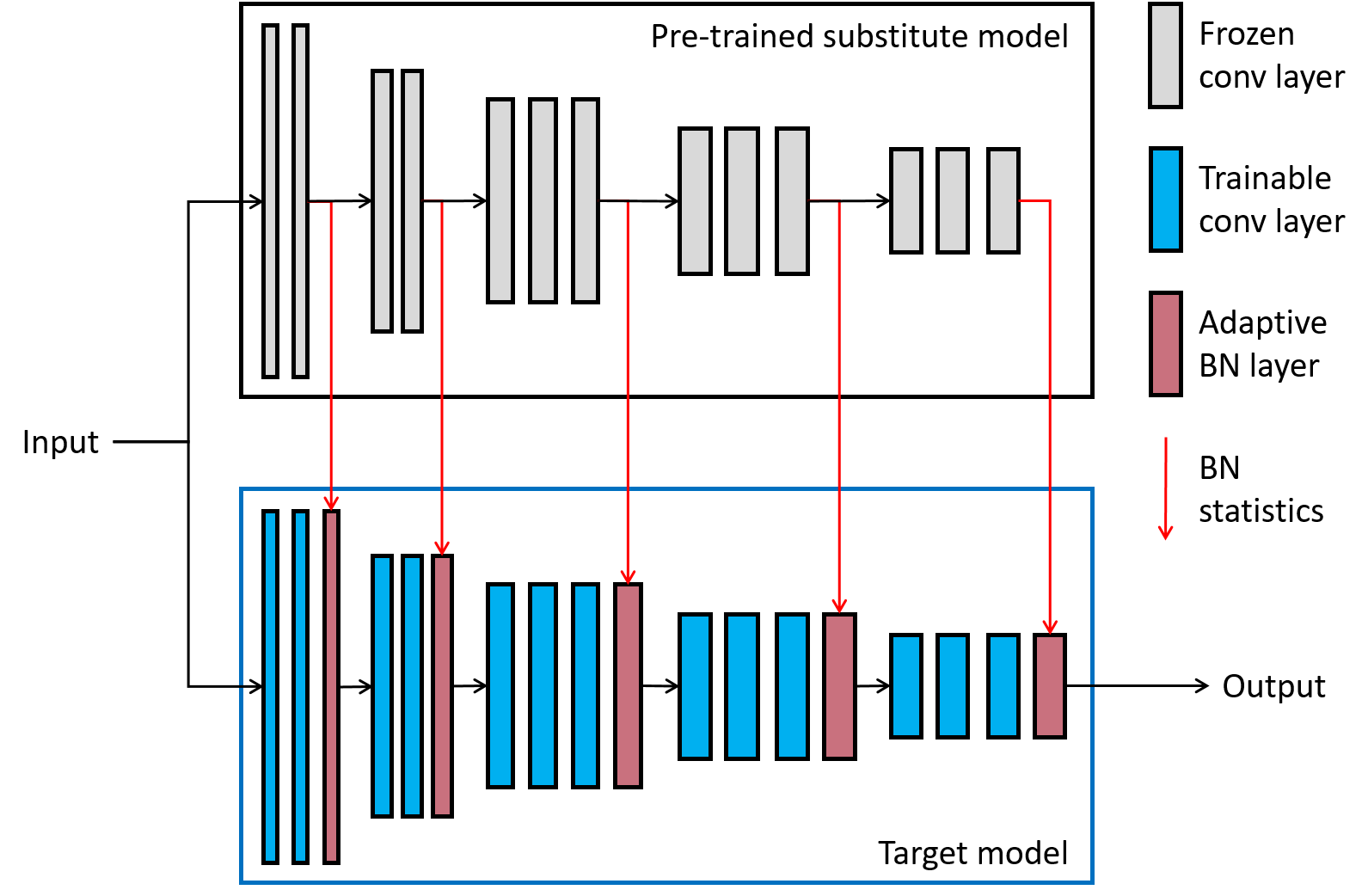}
		\caption{The proposed ABNN framework. ABNN employs a pre-trained and frozen substitute model to generate cleaner BN statistics, and sends them to the target model. The target model is exclusively trained on clean data and learns to align the substitute model's BN statistics.}
		\label{fig:abnn}
	\end{center}
\end{figure*}

To our purpose, we resort to an adaptive BN idea, inspired by Test-Time Adaptation (TTA) approaches \cite{li2017revisiting,valanarasu2023fly,wang2021tent}. We propose a novel adversarial defense accordingly, referred to as Adaptive Batch Normalization Network (ABNN). ABNN employs a pre-trained and frozen substitute model to generate cleaner BN statistics, and send them to the target model. The target model is exclusively trained on clean data and learns to align the substitute model's BN statistics. We extensively evaluate the proposed ABNN's robustness against both digital and physically realizable attack types in both image and video modalities. The experiments demonstrate that ABNN enhances adversarial robustness against both attack types in both modalities, with significantly lower training time complexity compared to AT-based approaches \cite{lo2021defending,lo2020overcomplete,madry2018towards}, and it can achieve higher clean data performance (see Figure~\ref{fig:cover}).

The key contributions of this work are summarized as follows:
\begin{itemize}
	\item We introduce a novel idea that uses test-time domain adaptation techniques to defend against adversarial examples.
	\item The proposed adversarial defense ABNN is a non-AT method that gets rid of the extremely time-consuming AT.
	\item Experiments show that ABNN can improve adversarial robustness against both digital and physically realizable attacks in both image and video modalities. Compared to AT-based approaches, it achieves higher clean data performance, better robustness generalization, and significantly lower training time complexity.
\end{itemize}

\section{Proposed Method}
As demonstrated in \cite{lo2022exploring,lo2021defending,xie2020adversarial,xie2020intriguing}, adversarial examples have different BN statistics from clean data. Such adversarial BN statistics cause accuracy drops. We see this property from a domain shift perspective and propose ABNN based on an adaptive BN idea. The framework of the proposed ABNN method is illustrated in Figure~\ref{fig:abnn}. It consists of a target model and a pre-trained substitute model. An input would pass through both models parallelly. We deploy our adaptive BN layer after each convolution block of the target model (e.g., after the conv1, conv2, conv3, conv4 and conv5 blocks of a ResNet \cite{he2016deep}).

\subsection{Adaptive Batch Normalization Layer}
Recall that BN \cite{ioffe2015batch} normalizes features to address the covariate shift problem, improving training efficiency and stability. A standard BN layer is defined as:
\begin{equation}
	z' = \gamma \bigg[ \frac{z - \mu(z)}{\sigma(z)} \bigg] + \beta,
\end{equation}
where $z$ is the input feature, $z'$ is the normalized output feature, $\{\mu(z), \sigma(z)\}$ denotes the BN statistics of mean and standard deviation, and $\{\gamma, \beta\}$ are trainable parameters for scaling and shifting, respectively. Let us consider that given an input sample, $z_t$ and $z_s$ represent its features extracted by the target and substitute models, respectively. Our adaptive BN layer receives $\{\mu(z_s), \sigma(z_s)\}$, the BN statistics estimated by the substitute model, then normalizes the target model's feature $z_t$ to $z'_t$ by:
\begin{equation}
	z'_t = \gamma_s \big[ \sigma(z_s) \bigg[ \frac{z_t - \mu(z_t)}{\sigma(z_t)} \bigg] + \mu(z_s) \big] + \beta_s.
\end{equation}
Inspired by \cite{valanarasu2023fly,valanarasu2023interactive}, we train the adaptive parameters $\{\gamma_s, \beta_s\}$ via the AdaIN \cite{huang2017arbitrary} encoding layer, i.e., $\{\gamma_s, \beta_s\} = AdaIN(z_s)$. $AdaIN$ encodes the substitute model's feature $z_s$ to derive the adaptive parameters used to align the BN statistics of the two models.

\subsection{Training and Inference}
We first pre-train the substitute model on one or multiple large-scale datasets (e.g., ImageNet \cite{deng2009imagenet}), where pre-training datasets are different from the target task dataset. This pre-training stage aims to learn a good feature extractor that can extract semantically meaningful features, thereby acquiring clean and high-quality BN statistics.

Next, we train the target model with the target task dataset. The target model learns its model parameters and the adaptive parameters $\{\gamma_s, \beta_s\}$. At this stage, the substitute model sends its corresponding BN statistics $\{\mu(z_s), \sigma(z_s)\}$ to the target model's adaptive BN layers, and the substitute model itself is frozen on training. Both the pre-training and target task training stages train on clean data exclusively without AT.

At inference time, the pipeline follows the same forward pass as the target task training stage. Under adversarial attacks, the target model's BN statistics are perturbed, resulting in indiscriminate features. In comparison, the substitute model's BN statistics are relatively unaffected even under white-box attacks, since the adversary focuses more on the target model to attack the target task. Moreover, the substitute model is pre-trained on large-scale datasets different from the target task dataset, making it harder for the adversary to transfer the attack to the substitute model. Our adaptive BN layer can adapt the substitute model's cleaner BN statistics to the target model, mitigating the adversarial effects in the target model's features.

\subsection{Discussion}
ABNN does not rely on AT, so it is much more training-efficient. We present an analysis of training complexity in Sec.~\ref{sec:time_analysis}. Besides, the pre-trained substitute model can be reused for any number of downstream target task models, saving additional training time. Avoiding AT enjoys better clean data performance as well. Furthermore, compared to image transformation-based defenses, the entire ABNN framework is fully differentiable and thus does not cause obfuscated gradients.

On the other hand, ABNN is partly related to the \textit{adaptive test-time defense} \cite{croce2022evaluating}, a defense category that delivers adaptive defense mechanisms at test time. However, these adaptive test-time defenses involve iterative optimization during inference, significantly increasing inference computation. In contrast, our ABNN does not have test time optimization. The only extra computation overhead is the substitute model, which is much lower than optimization. Their comparison is similar to the relation between standard TTA \cite{wang2021tent} and on-the-fly adaptation \cite{valanarasu2023fly}. In short, the proposed method takes adversarial robustness, clean data performance, training and inference efficiency into consideration, achieving a good balance among these aspects.

\section{Experiments} \label{sec:3}

\subsection{Experimental Setup}

\noindent \textbf{Datasets.}
We evaluate our method on both image and video modalities. For images, we use CIFAR-10 \cite{krizhevsky2009learning}, an image classification dataset that comprises 60,000 images with size $32 \times 32$ from 10 classes. For videos, we use UCF-101, an action recognition dataset that consists of 13,320 videos from 101 action classes. Following \cite{lo2021defending,lo2020overcomplete,wei2019sparse}, we resize the frame dimension to
$112 \times 112$ and uniformly sample each video into 40 frames

\noindent \textbf{Attack setting.}
We use the $L_\infty$-norm PGD \cite{madry2018towards} and the ROA \cite{Wu2020Defending} attacks to evaluate adversarial robustness. They cover the two most common attack types: digital attack and physically realizable attack, respectively. We set the PGD attack strength $\epsilon=8/255$ for images, $\epsilon=1/255$ for videos, and the number of attack iterations $t_{max}=5$ for both image and video data. For the ROA attack, we set the attacking area to $10\%$ of the input size. All the attacks are conducted under the white-box setting, i.e., we generate adversarial examples upon the entire framework, so the adversaries are fully aware of the defense. To reproduce AT-based defenses for comparison, we use the PGD attack with the same setting to do their AT.

\noindent \textbf{Implementation details.}
For image classification, we employ a ResNet-18 \cite{he2016deep} as the backbone network of the target model, and an ImageNet \cite{deng2009imagenet} pre-trained VGG-19 \cite{simonyan2015very} (with BN version) as the substitute model. For action recognition, we employ a 3D ResNeXt-101 \cite{hara3dcnns}, the 3D convolution version of ResNeXt-101 \cite{xie2017aggregated}, as the backbone network of the target model. We use a Kinetics-400 \cite{kay2017kinetics} pre-trained 3D ResNet-18 as the substitute model. At training time, the substitute model is frozen, and the target model is trained by the SGD optimizer. Experiments are implemented by PyTorch \cite{paszke2019pytorch} and performed on a single NVIDIA RTX 2080 Ti GPU.

\setlength{\tabcolsep}{10pt}
\begin{table}[t!]
	\begin{center}
		\caption{Evaluation results (\%) under the PGD attack on the CIFAR-10 dataset.}
		\label{table:cifar}
		\begin{tabular}{l | rr | r}
			\hline \noalign{\smallskip} \noalign{\smallskip}
			Method & Clean & PGD & Training cost \\
			\noalign{\smallskip} \hline \noalign{\smallskip}
			No Defense & 93.4 & 0.0 & 2N \\
			PGD-AT \cite{madry2018towards} & 83.3 & 51.6 & 12N \\
			ABNN (Ours) & 87.5 & 31.5 & 3N \\
			\noalign{\smallskip} \hline
		\end{tabular}
	\end{center}
\end{table}

\setlength{\tabcolsep}{9pt}
\begin{table}[t!]
	\begin{center}
		\caption{Evaluation results (\%) under the PGD attack on the UCF-101 dataset.}
		\label{table:ucf}
		\begin{tabular}{l | rr | r}
			\hline \noalign{\smallskip} \noalign{\smallskip}
			Method & Clean & PGD & Training cost \\
			\noalign{\smallskip} \hline \noalign{\smallskip}
			No Defense \cite{kinfu2022analysis} & 93.0 & 0.0 & 2N \\
			OUDefend \cite{lo2020overcomplete} & 62.0 & 58.6 & 24N \\
			ABNN (Ours) & 68.3 & 43.4 & 3N \\
			\noalign{\smallskip} \hline
		\end{tabular}
	\end{center}
\end{table}

\setlength{\tabcolsep}{9pt}
\begin{table}[t!]
	\begin{center}
		\caption{Evaluation results (\%) under the ROA attack on the UCF-101 dataset.}
		\label{table:roa}
		\begin{tabular}{l | rr | r}
			\hline \noalign{\smallskip} \noalign{\smallskip}
			Method & Clean & ROA & Training cost \\
			\noalign{\smallskip} \hline \noalign{\smallskip}
			No Defense \cite{kinfu2022analysis} & 93.0 & 7.0 & 2N \\
			OUDefend \cite{lo2020overcomplete} & 62.0 & 13.6 & 24N \\
			ABNN (Ours) & 68.3 & 24.4 & 3N \\
			\noalign{\smallskip} \hline
		\end{tabular}
	\end{center}
\end{table}

\subsection{Evaluation Results}

\begin{figure}[t!]
	\begin{center}
		\centering
		\includegraphics[width=0.48\textwidth]{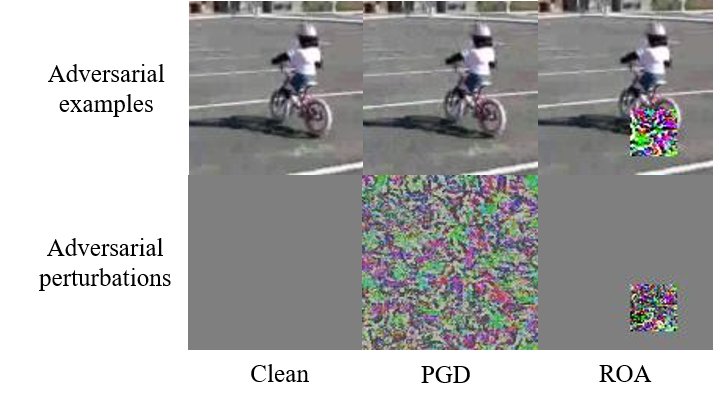}
		\caption{Examples of clean data, the PGD attack, and the ROA attack. The PGD attack is typically imperceptible to human eyes. The PGD perturbations are $15 \times$ magnified for better visualization. The ROA attack is like pasting a rectangular adversarial sticker on the input.}
		\label{fig:attacks}
	\end{center}
\end{figure}

\noindent \textbf{Image classification.}
Table~\ref{table:cifar} reports the robustness against PGD attack on image classification. We compare with No Defense, a baseline model without any defense mechanism, and PGD-AT \cite{madry2018towards}, a standard AT approach. We can observe that the proposed ABNN significantly improves robust accuracy from 0\% to 31.5\% on CIFAR-10 without using AT. On the other hand, ABNN only sacrifices 5.9\% clean accuracy, while PGD-AT sacrifices 10.1\%. Hence, although ABNN is not more robust than PGD-AT, it enjoys higher clean accuracy.

\noindent \textbf{Action recognition.}
Table~\ref{table:ucf} reports the robustness against PGD attack on action recognition. We compare with OUDefend \cite{lo2020overcomplete}, an AT-based state-of-the-art adversarial defense for video models. As can be seen, although ABNN does not include AT, its robust accuracy attains 43.4\%. This is close to the robust level achieved by the AT-based approach OUDefend. Moreover, ABNN obtains higher clean accuracy than OUDefend.

\noindent \textbf{Physically realizable attack.}
The physically realizable attack is a type of adversarial attack that the perturbations are printable and thus can be implemented in the physical space. ROA \cite{Wu2020Defending} is a widely used physically realizable attack. It performs $L_\infty$-norm PGD inside a fixed-size rectangle on an image or video frame. The perturbations can be viewed as pasting a rectangular adversarial sticker on the input (see Figure~\ref{fig:attacks}). In addition to the digital PGD attack, we further evaluate our method on the physically realizable ROA attack.

As shown in Table~\ref{table:roa}, the proposed ABNN achieves significantly higher robust accuracy against the ROA attack compared to OUDefend. This is because AT-based methods are typically robust only to the specific attack type that they are trained against. In other words, if they are optimized to defend against the PGD attack during AT, they may not be robust against other attack types like ROA. AT is known for this limited generalization. While several studies have developed generalizable AT methods \cite{lo2021defending}, they often lead to much higher training time complexity than regular AT. In contrast, we propose to employ the TTA idea to defend at test time without requiring any prior assumptions about the attack types. Therefore, our method enjoys high flexibility in defending against multiple types of attacks.

\subsection{Training Time Complexity} \label{sec:time_analysis}
The proposed method has significantly lower training time complexity than AT-based approaches. Here we make an analysis in detail. Let us set each network pass (i.e., a forward pass or a backward pass) to have $N$ computational complexity, and let us suppose that ABNN's target network and substitute network have the same complexity. Therefore, ABNN spends $2N$ on a forward pass, for it needs to pass through both networks. Since the substitute model is frozen during training, ABNN spends $N$ on a backward pass (passes through the target network only). The total complexity of a training step is:
\begin{equation}
	2N + N = 3N.
\end{equation}

PGD-AT requires generating multi-step adversarial examples for training. It spends $2N$ on each attack iteration (a forward pass plus a backward pass). It also spends $2N$ on training model parameters at each training step. Therefore, if the number of attack iterations is $t_{max}$, the complexity of a single complete training step would be:
\begin{equation}
	2N \cdot t_{max} + 2N = 2N(t_{max}+1).
\end{equation}
Hence, PGD-AT has:
\begin{equation}
	(t_{max}+1) \cdot 2N / 3N \simeq 0.67(t_{max}+1)
\end{equation}
times more training complexity than ABNN, which linearly increases along with the AT's $t_{max}$. OUDefend contains a feature-denoising network to remove adversaries in the feature space. Suppose both the feature-denoising network and the target network have $N$ computational complexity, the cost of each pass would be doubled. Hence, OUdefend's training complexity is:
\begin{equation}
	2 \cdot 2N(t_{max}+1) = 4N(t_{max}+1).
\end{equation}

We compare the training complexity of each method in Table~\ref{table:cifar}-\ref{table:roa} given $t_{max}=5$ in our experimental setting. ABNN has only $1/4$ and $1/8$ times training cost compared to PGD-AT and OUDefend, respectively. It clearly demonstrates that ABNN is much more efficient in terms of training computation.

\section{Conclusion}
In this paper, we propose a non-AT adversarial defense method, namely ABNN. With cleaner BN statistics sent from a pre-trained substitute mode, it is able to mitigate adversarial effects and thus improve robustness. Moreover, because ABNN avoids AT, it is not only much more training-efficient but also achieves better clean data performance and generalization. Adversarial robustness via domain adaptation ideas is less explored. We demonstrate that this is a promising direction and worth further exploration.

\section*{Acknowledgment}
This work was supported by the DARPA GARD Program HR001119S0026-GARD-FP-052.

{\small
	\bibliographystyle{ieee}
	\bibliography{thesis_cite}

\begin{thebibliography}{10}\itemsep=-1pt

\bibitem{obfuscated}
A.~Athalye, N.~Carlini, and D.~Wagner.
\newblock Obfuscated gradients give a false sense of security: circumventing
  defenses to adversarial examples.
\newblock In {\em International Conference on Machine learning (ICML)}, 2018.

\bibitem{chang2019domain}
W.-G. Chang, T.~You, S.~Seo, S.~Kwak, and B.~Han.
\newblock Domain-specific batch normalization for unsupervised domain
  adaptation.
\newblock In {\em IEEE/CVF Conference on Computer Vision and Pattern
  Recognition (CVPR)}, 2019.

\bibitem{chang2019all}
W.-L. Chang, H.-P. Wang, W.-H. Peng, and W.-C. Chiu.
\newblock All about structure: Adapting structural information across domains
  for boosting semantic segmentation.
\newblock In {\em IEEE/CVF Conference on Computer Vision and Pattern
  Recognition (CVPR)}, 2019.

\bibitem{croce2022evaluating}
F.~Croce, S.~Gowal, T.~Brunner, E.~Shelhamer, M.~Hein, and T.~Cemgil.
\newblock Evaluating the adversarial robustness of adaptive test-time defenses.
\newblock In {\em International Conference on Machine Learning (ICML)}, 2022.

\bibitem{deng2009imagenet}
J.~Deng, W.~Dong, R.~Socher, L.-J. Li, K.~Li, and L.~Fei-Fei.
\newblock Imagenet: A large-scale hierarchical image database.
\newblock In {\em IEEE/CVF Conference on Computer Vision and Pattern
  Recognition (CVPR)}, 2009.

\bibitem{ganin2015unsupervised}
Y.~Ganin and V.~Lempitsky.
\newblock Unsupervised domain adaptation by backpropagation.
\newblock In {\em International Conference on Machine learning (ICML)}, 2015.

\bibitem{goodfellow2015explaining}
I.~Goodfellow, J.~Shlens, and C.~Szegedy.
\newblock Explaining and harnessing adversarial examples.
\newblock In {\em International Conference on Learning Representations (ICLR)},
  2015.

\bibitem{guo2017countering}
C.~Guo, M.~Rana, M.~Cisse, and L.~Van Der~Maaten.
\newblock Countering adversarial images using input transformations.
\newblock In {\em International Conference on Learning Representations (ICLR)},
  2018.

\bibitem{hara3dcnns}
K.~Hara, H.~Kataoka, and Y.~Satoh.
\newblock Can spatiotemporal 3d cnns retrace the history of 2d cnns and
  imagenet?
\newblock In {\em IEEE/CVF Conference on Computer Vision and Pattern
  Recognition (CVPR)}, 2018.

\bibitem{he2016deep}
K.~He, X.~Zhang, S.~Ren, and J.~Sun.
\newblock Deep residual learning for image recognition.
\newblock In {\em IEEE/CVF Conference on Computer Vision and Pattern
  Recognition (CVPR)}, 2016.

\bibitem{huang2017arbitrary}
X.~Huang and S.~Belongie.
\newblock Arbitrary style transfer in real-time with adaptive instance
  normalization.
\newblock In {\em IEEE/CVF International Conference on Computer Vision (ICCV)},
  2017.

\bibitem{ioffe2015batch}
S.~Ioffe and C.~Szegedy.
\newblock Batch normalization: Accelerating deep network training by reducing
  internal covariate shift.
\newblock In {\em International Conference on Machine learning (ICML)}, 2015.

\bibitem{kay2017kinetics}
W.~Kay, J.~Carreira, K.~Simonyan, B.~Zhang, C.~Hillier, S.~Vijayanarasimhan,
  F.~Viola, T.~Green, T.~Back, P.~Natsev, et~al.
\newblock The kinetics human action video dataset.
\newblock In {\em arXiv preprint arXiv:1705.06950}, 2017.

\bibitem{kinfu2022analysis}
K.~A. Kinfu and R.~Vidal.
\newblock Analysis and extensions of adversarial training for video
  classification.
\newblock In {\em IEEE/CVF Conference on Computer Vision and Pattern
  Recognition Workshop (CVPRW)}, 2022.

\bibitem{krizhevsky2009learning}
A.~Krizhevsky.
\newblock Learning multiple layers of features from tiny images.
\newblock 2009.

\bibitem{li2017revisiting}
Y.~Li, N.~Wang, J.~Shi, J.~Liu, and X.~Hou.
\newblock Revisiting batch normalization for practical domain adaptation.
\newblock In {\em International Conference on Learning Representations Workshop
  (ICLRW)}, 2017.

\bibitem{liao2018defense}
F.~Liao, M.~Liang, Y.~Dong, T.~Pang, X.~Hu, and J.~Zhu.
\newblock Defense against adversarial attacks using high-level representation
  guided denoiser.
\newblock In {\em IEEE/CVF Conference on Computer Vision and Pattern
  Recognition (CVPR)}, 2018.

\bibitem{lo2023spatio}
S.-Y. Lo, P.~Oza, S.~Chennupati, A.~Galindo, and V.~M. Patel.
\newblock Spatio-temporal pixel-level contrastive learning-based source-free
  domain adaptation for video semantic segmentation.
\newblock In {\em IEEE/CVF Conference on Computer Vision and Pattern
  Recognition (CVPR)}, 2023.

\bibitem{lo2021defending}
S.-Y. Lo and V.~M. Patel.
\newblock Defending against multiple and unforeseen adversarial videos.
\newblock In {\em IEEE Transactions on Image Processing (T-IP)}, 2021.

\bibitem{lo2022exploring}
S.-Y. Lo and V.~M. Patel.
\newblock Exploring adversarially robust training for unsupervised domain
  adaptation.
\newblock In {\em Asian Conference on Computer Vision (ACCV)}, 2022.

\bibitem{lo2020overcomplete}
S.-Y. Lo, J.~M.~J. Valanarasu, and V.~M. Patel.
\newblock Overcomplete representations against adversarial videos.
\newblock In {\em IEEE International Conference on Image Processing (ICIP)},
  2021.

\bibitem{lo2022learning}
S.-Y. Lo, W.~Wang, J.~Thomas, J.~Zheng, V.~M. Patel, and C.-H. Kuo.
\newblock Learning feature decomposition for domain adaptive monocular depth
  estimation.
\newblock In {\em IEEE/RSJ International Conference on Intelligent Robots and
  Systems (IROS)}, 2022.

\bibitem{madry2018towards}
A.~Madry, A.~Makelov, L.~Schmidt, D.~Tsipras, and A.~Vladu.
\newblock Towards deep learning models resistant to adversarial attacks.
\newblock In {\em International Conference on Learning Representations (ICLR)},
  2018.

\bibitem{paszke2019pytorch}
A.~Paszke, S.~Gross, F.~Massa, A.~Lerer, J.~Bradbury, G.~Chanan, T.~Killeen,
  Z.~Lin, N.~Gimelshein, L.~Antiga, et~al.
\newblock Pytorch: An imperative style, high-performance deep learning library.
\newblock In {\em Conference on Neural Information Processing Systems
  (NeurIPS)}, 2019.

\bibitem{simonyan2015very}
K.~Simonyan and A.~Zisserman.
\newblock Very deep convolutional networks for large-scale image recognition.
\newblock In {\em International Conference on Learning Representations (ICLR)},
  2015.

\bibitem{szegedy2014intriguing}
C.~Szegedy, W.~Zaremba, I.~Sutskever, J.~Bruna, D.~Erhan, I.~Goodfellow, and
  R.~Fergus.
\newblock Intriguing properties of neural networks.
\newblock In {\em International Conference on Learning Representations (ICLR)},
  2014.

\bibitem{tsipras2018robustness}
D.~Tsipras, S.~Santurkar, L.~Engstrom, A.~Turner, and A.~Madry.
\newblock Robustness may be at odds with accuracy.
\newblock In {\em International Conference on Learning Representations (ICLR)},
  2019.

\bibitem{valanarasu2023fly}
J.~M.~J. Valanarasu, P.~Guo, V.~VS, and V.~M. Patel.
\newblock On-the-fly test-time adaptation for medical image segmentation.
\newblock In {\em Medical Imaging with Deep Learning (MIDL)}, 2023.

\bibitem{valanarasu2023interactive}
J.~M.~J. Valanarasu, H.~Zhang, J.~Zhang, Y.~Wang, Z.~Lin, J.~Echevarria, Y.~Ma,
  Z.~Wei, K.~Sunkavalli, and V.~M. Patel.
\newblock Interactive portrait harmonization.
\newblock In {\em International Conference on Learning Representations (ICLR)},
  2023.

\bibitem{wang2021tent}
D.~Wang, E.~Shelhamer, S.~Liu, B.~Olshausen, and T.~Darrell.
\newblock Tent: Fully test-time adaptation by entropy minimization.
\newblock In {\em International Conference on Learning Representations (ICLR)},
  2021.

\bibitem{wei2019sparse}
X.~Wei, J.~Zhu, S.~Yuan, and H.~Su.
\newblock Sparse adversarial perturbations for videos.
\newblock In {\em AAAI Conference on Artificial Intelligence (AAAI)}, 2019.

\bibitem{Wu2020Defending}
T.~Wu, L.~Tong, and Y.~Vorobeychik.
\newblock Defending against physically realizable attacks on image
  classification.
\newblock In {\em International Conference on Learning Representations (ICLR)},
  2020.

\bibitem{xie2020adversarial}
C.~Xie, M.~Tan, B.~Gong, J.~Wang, A.~Yuille, and Q.~V. Le.
\newblock Adversarial examples improve image recognition.
\newblock In {\em IEEE/CVF Conference on Computer Vision and Pattern
  Recognition (CVPR)}, 2020.

\bibitem{xie2020intriguing}
C.~Xie and A.~Yuille.
\newblock Intriguing properties of sdversarial training at scale.
\newblock In {\em International Conference on Learning Representations (ICLR)},
  2020.

\bibitem{xie2017aggregated}
S.~Xie, R.~Girshick, P.~Doll{\'a}r, Z.~Tu, and K.~He.
\newblock Aggregated residual transformations for deep neural networks.
\newblock In {\em IEEE/CVF Conference on Computer Vision and Pattern
  Recognition (CVPR)}, 2017.

\bibitem{xu2017feature}
W.~Xu, D.~Evans, and Y.~Qi.
\newblock Feature squeezing: Detecting adversarial examples in deep neural
  networks.
\newblock In {\em Network and Distributed System Security Symposium (NDSS)},
  2018.

\bibitem{yang2021adversarial}
S.~Yang, T.~Guo, Y.~Wang, and C.~Xu.
\newblock Adversarial robustness through disentangled representations.
\newblock In {\em AAAI Conference on Artificial Intelligence (AAAI)}, 2021.

\bibitem{zhang2019theoretically}
H.~Zhang, Y.~Yu, J.~Jiao, E.~P. Xing, L.~E. Ghaoui, and M.~I. Jordan.
\newblock Theoretically principled trade-off between robustness and accuracy.
\newblock In {\em International Conference on Machine learning (ICML)}, 2019.

\end{thebibliography}
}

\end{document}